\documentclass{article}
\PassOptionsToPackage{numbers, compress}{natbib}
\usepackage[final,main]{neurips_2026}

\usepackage[utf8]{inputenc}
\usepackage[T1]{fontenc}
\usepackage{hyperref}
\usepackage{url}
\usepackage{graphicx}
\usepackage{booktabs}
\usepackage{amsfonts}
\usepackage{amsmath,amssymb}
\usepackage{nicefrac}
\usepackage{microtype}
\usepackage{xcolor}
\usepackage{algorithm}
\usepackage{algpseudocode}
\usepackage{booktabs}
\usepackage{multirow}
\usepackage{adjustbox}
\usepackage{wrapfig}
\usepackage{tikz}
\usepackage{pifont}
\newcommand{\cmark}{\ding{51}}

\title{ForceLDM: A Force-aware Latent Dynamics Model for Contact-Rich Manipulation} 

\author{%
  \normalfont\bfseries
  Peiyuan Tang\textsuperscript{1,4*} \quad
  Yao Li\textsuperscript{2,3*} \quad
  Xiaodong Zhang\textsuperscript{5\textdagger} \quad
  Wuyang Zhang\textsuperscript{2,3,4} \\ \bfseries
  Qin Xia\textsuperscript{1} \quad
  Yanyong Zhang\textsuperscript{2} \quad
  Zijiang James Yang\textsuperscript{5,6\textdagger} \\[3pt]
  \normalfont
  \textsuperscript{1}School of Computer Science and Technology, Xi'an Jiaotong University \\
  \textsuperscript{2}School of Artificial Intelligence and Data Science, University of Science and Technology of China \\
  \textsuperscript{3}Suzhou Institute for Advanced Research, University of Science and Technology of China \\
  \textsuperscript{4}Suzhou Artificial Intelligence Laboratory \\
  \textsuperscript{5}School of Computer Science and Technology, University of Science and Technology of China \\
  \textsuperscript{6}Synkrotron, Inc. \\[2pt]
  \textsuperscript{*}Equal contribution \quad
  \textsuperscript{\textdagger}Corresponding authors.
}

\begin{document}

\maketitle
 
\begin{abstract}
Contact-rich manipulation requires fine-grained control and complex physical interactions with the environment, posing significant challenges for robot policy learning. 
Vision-only policies struggle with such tasks, as contact states and interaction forces are difficult to infer from visual observations alone.
Existing methods mitigate this issue by incorporating force or tactile sensing. 
However, they often treat these signals as auxiliary observations without fully exploiting their role in modeling future interaction dynamics.
To address this, we propose ForceLDM, a force-aware latent dynamics model that learns future contact and motion dynamics to guide action generation.
Rather than directly predicting future observations, which can entangle task-relevant dynamics with irrelevant sensory details, ForceLDM learns a compact representation of future dynamics through knowledge distillation.
Specifically, a teacher network with privileged future force/torque and optical flow signals extracts future dynamic features, which are then distilled into a student network trained only on current observations.
This enables the student model to reason about upcoming contacts and physical dynamics without relying on future sensory inputs at deployment.
To improve robustness and generalization, we further introduce a curriculum-based progressive noise injection strategy to mitigate over-reliance on future features.
Experiments on five real-world contact-rich manipulation tasks demonstrate that ForceLDM outperforms state-of-the-art baselines, generalizes effectively to novel objects, and remains robust to environmental perturbations.
Project page and code are available at \url{https://tpy001.github.io/ForceLDM/}.

\end{abstract}

\section{Introduction}
\label{sec:intro}
Humans are naturally capable of performing dexterous manipulation such as USB insertion and component assembly. However, these simple daily tasks remain highly challenging for robots. 
Contact-rich manipulation, particularly high-precision insertion, requires not only accurate motion
planning and execution, but also the ability to perceive and respond to contact forces in real time~\citep{Foar,rdp,favla}.
Although vision-based policies have achieved remarkable progress in general-purpose robotic manipulation~\citep{OpenVLA,pi0,pi05,GR00TNA}, they still struggle with
contact-rich tasks, as contact states and force magnitudes are often difficult to infer reliably from vision alone. This limitation has motivated recent studies~\citep{ForceVLA,ForceVLA2,Forcemimic,TAVLA,rdp} to incorporate physical feedback, such as force or torque sensing, to improve contact-rich manipulation performance.

Recent studies on force-aware contact-rich manipulation have primarily focused on designing more effective multimodal architectures to fully exploit force/torque feedback. For example, ForceVLA~\citep{ForceVLA} employs a Mixture-of-Experts (MoE) architecture with modality-specific experts to adaptively fuse visual and force features. TA-VLA~\citep{TAVLA} investigates joint torque fusion in VLA models and shows that injecting torque signals into the action generation module improves performance.
Other works~\citep{rdp,Omnivta} explore fast-slow dual-system architectures that combine low-frequency vision with high-frequency force feedback for responsive control.

Despite these advances, most work still treats force/torque signals as additional inputs without modeling future interaction dynamics.
This leads to two key limitations.
First, high-dimensional visual observations provide rich global semantics, whereas low-dimensional force/torque signals are spatiotemporally sparse and only effective during the contact stage.
Consequently, models tend to rely on visual cues while failing to fully exploit force feedback~\citep{FACT,PolicyConsensus,VT-WAM}.
Second, most force-aware VLAs  learn only a mapping from current observations to actions, which may lead to overfitting to expert behaviors and limit generalization to unseen environments.

\begin{figure}[t]
    \centering 
    \includegraphics[width=0.98\linewidth]{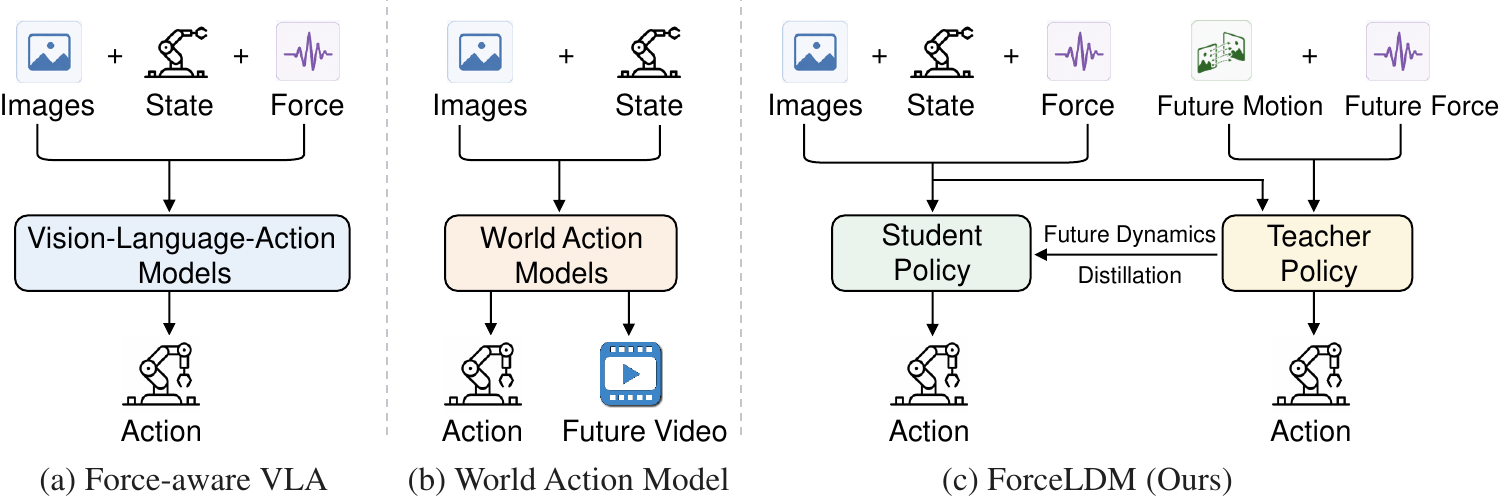}
    \caption{
    Comparison with prior methods.
    (a) Force-aware VLAs incorporate force feedback but lack explicit modeling of future interaction dynamics.
    (b) Video-based WAMs model future dynamics through visual prediction, but often lack force/tactile feedback and may fit task-irrelevant visual details.
    (c) ForceLDM learns compact, action-relevant future dynamics by distilling future motion and force cues from a privileged teascher.
    }
    \label{fig:concept}
\end{figure}
\vspace{-2mm}

A natural approach to addressing these limitations is to incorporate world models into VLAs, learning dynamics modeling through future observation prediction~\citep{VPP,Seer,Dreamvla}.
However, we find that integrating this idea into force-aware VLAs is non-trivial.
On the one hand, most existing world models are based on video generation models. They can generate visually realistic videos, but may fail to capture physically consistent contact dynamics. Fine-tuning these models with force observations requires substantial computational resources and large amounts of synchronized multimodal data.
On the other hand, predicting future observations often entangles task-relevant dynamics with task-irrelevant details, such as backgrounds, object appearances, and sensor noise. This may lead models to focus on irrelevant information rather than dynamics that are critical for control.

To address these challenges, we propose \textbf{ForceLDM}, a force-aware latent dynamics model for contact-rich manipulation. 
As illustrated in Figure~\ref{fig:concept}, ForceLDM uses optical flow and force/torque signals as dynamics representations, providing direct information about motion and contact.
To effectively model these dynamics, we propose a Future-Dynamics Distillation framework that distills future dynamics from a teacher to a student model.
Specifically, a privileged teacher model extracts compact dynamics features from future observations by predicting actions.
This encourages the teacher to selectively retain information that is most useful for control.
A student policy is then trained to match these latent representations using a set of learnable query tokens.
This allows the student to imagine future dynamics in the teacher's latent space without reconstructing future observations in full detail.
To further improve generalization, we introduce curriculum-based Progressive Noise Injection, which gradually increases noise on the student's query tokens to reduce overfitting.
Together, these designs enable ForceLDM to learn control-relevant future dynamics and improve generalization in contact-rich manipulation.

We evaluate ForceLDM on five real-world contact-rich manipulation tasks under both in-distribution and out-of-distribution settings. 
ForceLDM achieves stronger task performance and better generalization to unseen visual and physical variations than existing VLA and force-aware baselines. 
Further ablations and representation analyses show that the student queries encode predictive information about future force and motion dynamics, indicating that ForceLDM implicitly learns physical interaction dynamics rather than merely fitting expert actions. 
Our contributions are summarized as follows:
\begin{itemize}
    \item We propose a future dynamics representation where force/torque reflects contact states and optical flow encodes motion dynamics, overcoming the limitations of purely visual representations in contact-rich manipulation.
    \item We introduce ForceLDM, a force-aware latent dynamics model that enables effective dynamics modeling from current observations, improving policy robustness and generalization to unseen objects and environmental variations.
    \item We empirically validate ForceLDM on five real-world contact-rich manipulation tasks, demonstrating its effectiveness in improving task performance, out-of-distribution generalization, and implicit dynamics learning.
\end{itemize}


\section{Related Work}
\textbf{Vision-Language-Action Models.}
Recent studies have leveraged pretrained vision-language models to build general-purpose robotic policies that follow open-ended instructions and perform diverse manipulation tasks~\citep{RT2,pi0,RDT1B,GR00TNA}.
These methods encode visual observations and language instructions into multimodal representations and map them to robot actions. Existing approaches adapt VLMs into robot policies by either autoregressively predicting discretized action tokens~\citep{OpenVLA,pi0fast} or directly modeling continuous action distributions via flow matching~\citep{pi0,GR00TNA}. 
Building on these foundations, recent studies further improve VLA models by enhancing high-level reasoning and planning~\citep{Onetwovla,LaST}, strengthening 3D spatial perception~\citep{Spatialvla,Spatial-forcing}, incorporating memory mechanisms~\citep{Memoryvla,Mem}, and developing dual-system architectures~\citep{Fast-in-slow,Hume}. 
Despite these advances, most VLA policies rely primarily on visual observations and robot states, which provide limited information about contact events, force directions and friction.
To bridge this gap, we introduce ForceLDM, which fuses force/torque sensing with vision-language representations to improve precision and compliance in contact-rich manipulation.


\textbf{Contact-Rich Manipulation with Physical Feedback.} 
Several approaches have explored incorporating physical feedback into robotic policies. Forcemimic~\citep{Forcemimic} and FoAR~\citep{Foar} introduce force-aware policies by integrating 6-DoF force/torque measurements, enabling robots to better perceive interaction dynamics. 
Other works~\citep{TAVLA,Factr2} exploit proprioceptive signals, such as joint efforts, to estimate external contact states without additional force sensors. Beyond force-based sensing, tactile-based approaches leverage high-resolution tactile images to capture localized contact geometry and surface interactions~\citep{3d-vitac,Tla,Tactile-VLA,Vla-touch}. 
To fuse these physical modalities with vision-language features, existing methods adopt feature concatenation~\citep{Vtla,TAVLA,Tla}, cross-attention~\citep{ViTacFormer,TouchInTheWild,ManipForce}, gating mechanisms~\citep{Foar,TacVLA,Omnivta,When2Feel}, and Mixture-of-Experts architectures~\citep{MOE,ForceVLA,ForceVLA2}. 
Building on this, recent works~\citep{TacForeSight,Omnivta,DreamTac,VT-WAM} incorporate world models to reason about future physical interactions by jointly predict future force or tactile signals and future videos. 
However, this multimodal prediction paradigm requires large-scale synchronized visual and force/tactile data and incurs substantial computational overhead, leading to high inference latency.
In contrast, our method distills task-relevant future force and motion signals into the policy, enabling efficient inference by avoiding video generation.

\textbf{World Modeling for VLA Policies.}
VLA policies learn a direct mapping from current visual-language observations to robot actions~\citep{RT2,OpenVLA,pi0}, whereas world models learn how future world states evolve through interaction~\citep{Dreamer,UWM,DreamZero}. Incorporating world modeling into VLA policies augments action supervision with future-state prediction, allowing action generation to account for future consequences rather than expert actions alone~\citep{VPP,finn2017deep,Dreamvla,Seer}. Early world-action models realize this idea through future image or video prediction~\citep{UWM,DreamZero}, whereas latent world models replace expensive pixel-space rollouts with compact, task-relevant predictive representations. In contrast to approaches that derive action-oriented predictive representations primarily from future visual observations~\citep{Flare,WorldGuidance,Being-H07}, ForceLDM distills privileged future force/torque and optical-flow dynamics into the policy to capture contact physics that may be visually subtle or occluded, without requiring future inputs or explicit rollouts at inference time.

\section{Method}
\label{sec:method}
We introduce \textbf{ForceLDM}, a latent dynamics learning framework that distills future force and motion dynamics into vision-language-action models, enabling them to better model physical interactions. In this section, we begin with the problem formulation, followed by an overview of the framework, and then describe the training and inference strategies.

\subsection{Problem Formulation}
\label{sec:problem_formulation}
We formulate contact-rich manipulation as a policy learning problem.
At each time step $t$, the policy $\pi_{\theta}$ takes the multimodal observation $\mathbf{O}_t = (I_t, s_t, F_t, l)$ as input, and outputs an action chunk $\mathbf{A}_t=(\mathbf{a}_t,\ldots,\mathbf{a}_{t+H-1})$ consisting of $H$ actions to be executed sequentially:
\begin{equation}
    \mathbf{A}_t = \pi_\theta(I_t, s_t, F_t, l),
\end{equation}
where $I_t$ denotes multi-view images, $s_t$ denotes the robot state (end-effector pose or joint positions), $F_t$ denotes 6D wrench measurements, and $l$ denotes the language instruction.
To train the policy, we collect a dataset consisting of $N$ expert demonstrations, denoted as $\mathcal{D} = \{\tau_i\}_{i=1}^N$, where each trajectory is defined as $\tau_i = \{(\mathbf{O}_t^i, \mathbf{a}_t^i)\}_{t=1}^{T_i}$. 
The policy  $\pi_{\theta}$ is optimized via behavior cloning~\citep{BehaviouralCloning,BehaviouralCloning2} by minimizing the discrepancy between predicted actions and expert demonstrated actions:
\begin{equation}
    \mathcal{L}_{\mathrm{BC}}(\theta)
    =
    \mathbb{E}_{(\mathbf{O}_t, \mathbf{A}_t) \sim \mathcal{D}}
    \left[
        \ell_{\mathrm{act}}\bigl(\mathbf{A}_t, \pi_\theta(\mathbf{O}_t)\bigr)
    \right],
\end{equation}
where $\ell_{\mathrm{act}}$ denotes the action prediction loss, commonly defined as the mean absolute error (MAE) or mean squared error (MSE).

\subsection{Overall Architecture}
\label{sec:overall_architecture}
\begin{figure}[t]
\centering
\includegraphics[width=\linewidth]{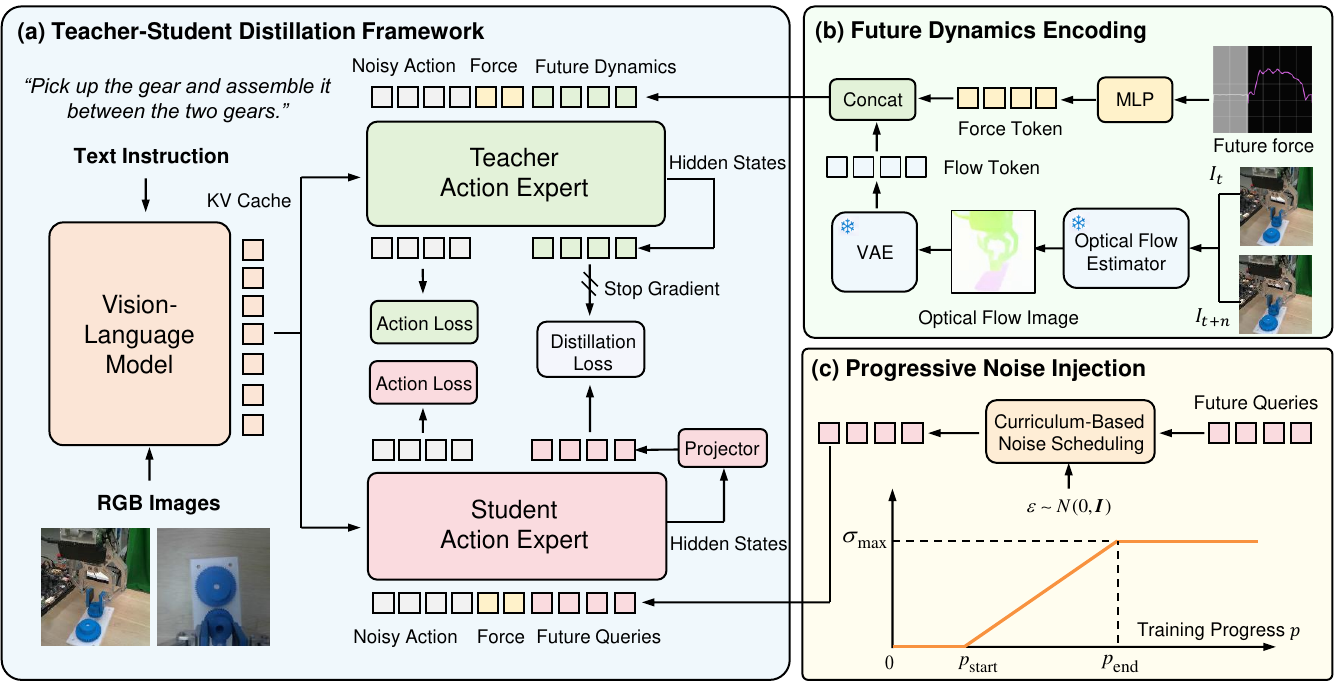}
\caption{
Overview of ForceLDM.
(a) Both teacher and student generate actions from current multimodal observations and language instructions. The teacher additionally receives future force and motion tokens, and its future-aware representations are distilled into the student's future queries.
(b) The future dynamics encoder converts future force/torque and optical flow into compact force and motion tokens.
(c) Progressive noise injection perturbs the student's future queries with scheduled noise to improve generalization.
}
\label{fig:framework}
\end{figure}

We aim to improve contact-rich manipulation by learning future interaction dynamics. Instead of predicting all details of future observations, we adopt a teacher-student distillation framework to focus on information relevant to robot control.
As illustrated in Figure~\ref{fig:framework}, ForceLDM consists of a shared VLM backbone, a future-dynamics encoder, and two action experts.
The VLM backbone encodes current images and language instructions to provide scene context.
Robot states and force history are embedded as tokens and fed into the action experts.
During training, the teacher takes future force and optical flow as inputs and learns compact representations of physical interactions through action prediction.
These representations are then distilled into the student through representation alignment, allowing it to infer future interactions from current observations.
To improve generalization, we further introduce progressive noise injection, which gradually adds noise to the student's future queries to prevent overfitting.
At inference time, only the student policy is retained for deployment.

\vspace{-2mm}

\subsection{Future Dynamics Encoding}
\label{sec:future_dynamics_encoding}
Prior approaches~\citep{VPP,Seer,zeng2024learning} often predict future visual observations, but RGB images are sensitive to changes in lighting and appearance and cannot fully capture physical interactions such as contact and resistance.
We therefore use future force/torque measurements and optical flow as complementary sources of information.
Force/torque measurements provide direct feedback on physical contact, while optical flow captures the motion of the robot and objects.
Their combination provides more informative cues for robot control than RGB frames alone.

\textbf{Future Force Encoding.}
We use an MLP to encode the 6D force/torque measurements over the next $H$ steps into a compact force token:
\begin{equation}
    \mathbf{f}_t = E_f\left(\mathbf{F}_{t:t+H}\right).
\end{equation}
This token captures changes in contact forces over time and provides information about the robot's physical interactions with its environment.

\textbf{Future Motion Encoding.}
Optical flow provides a dense representation of scene motion by describing the pixel-wise displacement between two frames. Given the current image $I_t$ and a future image $I_{t+H}$, we use a pretrained optical-flow estimation model~\citep{Raft} to obtain a dense motion field. For each pixel $p$, the corresponding flow vector $\mathbf{u}(p)=(u_x(p),u_y(p))$ represents its horizontal and vertical displacement in image space, thereby capturing the image-space motion of the robot and manipulated objects.
To align the optical flow with the image representation used in our framework, we colorize the two-channel flow field into a three-channel motion image $\mathbf{M}_{t\rightarrow t+H}$. Following FlowVLA~\citep{flowvla}, we map flow direction to hue and use the normalized, clipped magnitude to control the strength of the resulting direction color against a white background. We then compress $\mathbf{M}_{t\rightarrow t+H}$ using a pretrained image VAE~\citep{Sdxl} and employ $N_m$ learnable motion queries $Q_m$ to aggregate the encoded features into compact motion tokens:
\begin{equation}
\begin{aligned}
    \mathbf{Z}_m
    &= E_m\left(\mathbf{M}_{t\rightarrow t+H}\right),\\
    \mathbf{m}_t
    &= \operatorname{CrossAttn}\left(\mathbf{Q}_m,\mathbf{Z}_m\right).
\end{aligned}
\end{equation}
where $E_m$ denotes the pretrained VAE model.

\subsection{Online Future-Dynamics Distillation}
\label{sec:future_dynamics_distillation}
Directly predicting future observations may force the model to learn details irrelevant to control. It may also lead to overfitting to sensor noise in force/torque measurements and estimation errors in optical flow. To avoid these issues, we learn future dynamics through latent representation alignment, without explicitly reconstructing future observations in detail.

The student uses $N_m+1$ learnable queries
$\mathbf{Q}\in\mathbb{R}^{(N_m+1)\times D}$,
comprising one force query and $N_m$ motion queries,
to predict the teacher's future-aware intermediate representations.
We perform multi-layer distillation over a set of selected layers $K$.
At each layer $l\in K$, let
$\smash{\mathbf{h}_f^{(l)}}$ and $\smash{\mathbf{y}_f^{(l)}}$
denote the student and teacher force representations,
while $\smash{\mathbf{h}_m^{(l,i)}}$ and
$\smash{\mathbf{y}_m^{(l,i)}}$ denote their $i$-th
motion representations.
The distillation loss aligns both modalities using cosine similarity:
\begin{equation}
\resizebox{0.95\linewidth}{!}{$\displaystyle
\mathcal{L}_{\mathrm{distill}}
= \frac{1}{|K|}\sum_{l\in K}
\left[
1-\cos\left(
g_f(\mathbf{h}_f^{(l)}),
\mathrm{sg}(\mathbf{y}_f^{(l)})
\right)
+
\frac{1}{N_m}\sum_{i=1}^{N_m}
\left[
1-\cos\left(
g_m(\mathbf{h}_m^{(l,i)}),
\mathrm{sg}(\mathbf{y}_m^{(l,i)})
\right)
\right]
\right].
$}
\end{equation}
Here, $g_f$ and $g_m$ denote the projection heads for force
and motion representations, respectively, while
$\mathrm{sg}(\cdot)$ denotes the stop-gradient operation
that prevents gradients from propagating to the teacher.


\subsection{Training and Inference}
\label{sec:training_strategy}
\textbf{Progressive Noise Injection.}
\label{sec:noise_injection}
Future interaction dynamics are inherently uncertain, and limited training data can cause the student to overfit.
To improve generalization, we introduce progressive noise injection by adding Gaussian noise to the student queries.
Let $\mathbf{Q}$ and $\widetilde{\mathbf{Q}}$ be the original and perturbed student queries, respectively, $p\in[0,1]$ be the normalized training progress, and $\sigma(p)$ be the corresponding noise scale.
We perturb the queries with RMS-scaled Gaussian noise:
\begin{equation}
\widetilde{\mathbf{Q}}
=
\mathbf{Q}
+
\sigma(p) \cdot \operatorname{RMS}(\mathbf{Q}) \cdot \boldsymbol{\epsilon},
\qquad
\boldsymbol{\epsilon}\sim\mathcal{N}(\mathbf{0},\mathbf{I}),
\end{equation}
where $\operatorname{RMS}(\mathbf{Q})$ denotes the root mean square of the query values. The noise scale $\sigma(p)$ remains zero until $p_{\mathrm{start}}$, then increases linearly to $\sigma_{\max}$ at $p_{\mathrm{end}}$, and remains at $\sigma_{\max}$ thereafter. This schedule first allows the student to learn meaningful representations through alignment and then improves its robustness to query perturbations.

\textbf{Action Generation via Flow Matching.}
\label{sec:training_objective}
We train both the teacher and student policies using flow matching.
Given a ground-truth action chunk $\mathbf{a}$, we sample $\boldsymbol{\epsilon}\sim\mathcal{N}(\mathbf{0},\mathbf{I})$ and $\tau\sim\mathcal{U}(0,1)$, construct $\mathbf{a}_{\tau}=(1-\tau)\boldsymbol{\epsilon}+\tau\mathbf{a}$, and define the target velocity as $\mathbf{v}^*=\mathbf{a}-\boldsymbol{\epsilon}$.
Let $c_S$ and $c_T$ represent the student and teacher conditioning information, respectively. Their flow matching losses are:
\begin{equation}
\mathcal{L}_{S}=\mathbb{E}\!\left[\left\|\mathbf{v}_{S}(\mathbf{a}_{\tau},\tau,c_S)-\mathbf{v}^*\right\|_{2}^{2}\right],
\qquad
\mathcal{L}_{T}=\mathbb{E}\!\left[\left\|\mathbf{v}_{T}(\mathbf{a}_{\tau},\tau,c_T)-\mathbf{v}^*\right\|_{2}^{2}\right].
\end{equation}
Both policies use the same noise sample and interpolation time for consistent velocity prediction.
The final training objective is:
\begin{equation}
\mathcal{L}_{\mathrm{total}}
=
\mathcal{L}_{S}
+
\mathcal{L}_{T}
+
\lambda\mathcal{L}_{\mathrm{distill}},
\end{equation}
where $\lambda=0.5$ controls the weight of the distillation loss.

\textbf{Inference.}
During inference, only the student policy is retained, while the teacher and future-dynamics encoder are removed.
The student generates actions using current observations without requiring future force or motion information.

\section{Experiments}
\label{sec:experiments}

In this section, we evaluate our method on contact-rich manipulation tasks through extensive experiments. We first report results on five real-world tasks and assess out-of-distribution generalization. We then conduct ablation studies on future prediction targets, model architecture, and training strategies, followed by visualizations of the learned latent dynamics.

\subsection{Experimental Setup}
\label{sec:real_world}

\paragraph{Hardware Platform and Data Collection}
Our main experiments use a 7-DoF xArm 7 manipulator equipped with wrist-mounted and fixed third-person cameras and a 6D force/torque sensor.
We collect successful demonstrations through 3D SpaceMouse teleoperation at 30\,Hz, recording synchronized visual observations, robot states, force/torque measurements, and actions for policy training.
Detailed hardware configurations and data collection procedures are provided in Appendix~\ref{sec:data_collection}.

\begin{figure}[t]
    \centering
    \includegraphics[width=\linewidth]{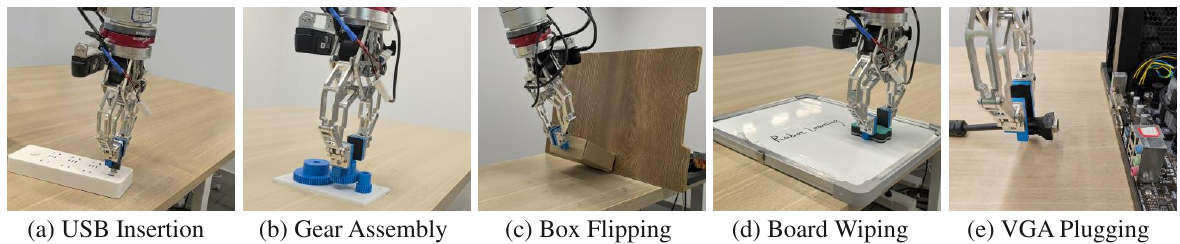}
    \vspace{-5mm}
    \caption{
    Five real-world contact-rich manipulation tasks used in our experiments.
    }
    \label{fig:real_tasks}
    \vspace{-2mm}
\end{figure}

\paragraph{Tasks and Evaluation Metrics.}
We evaluate ForceLDM on five real-world contact-rich manipulation tasks: USB Insertion, Gear Assembly, Box Flipping, Board Wiping, and VGA Plugging, as shown in Figure~\ref{fig:real_tasks}.
For each task, we collect 80--100 successful demonstration trajectories and evaluate each method over 20--30 real-world rollouts.
We report task success rate over repeated real-world trials as the primary evaluation metric.
Detailed task descriptions, success criteria, and evaluation protocols are provided in Appendix~\ref{sec:data_collection}.

\paragraph{Compared Baselines.}
We compare our method against several strong baselines.
\textbf{\(\boldsymbol{\pi_0}\)~\citep{pi0} w/o Force} is a flow-matching-based Vision-Language-Action model pretrained on large-scale robot data.
\textbf{\(\boldsymbol{\pi_0}\) w/ Force} is a force-augmented variant of \(\pi_0\) that converts force measurements into tokens and concatenates them with the input sequence.
\textbf{TA-VLA}~\citep{TAVLA} is a force-aware VLA model that introduces an auxiliary future-force prediction objective.
\textbf{ForceVLA}~\citep{ForceVLA} is a force-aware VLA model that adopts a Mixture-of-Experts architecture~\citep{MOE} to model visual and force modalities with specialized experts.
\textbf{Seer~\citep{Seer} w/ Force} augments Seer with force input and uses future RGB reconstruction as auxiliary supervision.
For a fair comparison, all methods are fully fine-tuned on the same collected dataset using the same training configuration. Detailed baseline implementations and training settings are provided in Appendix~\ref{app:baseline_details}.

\paragraph{Implementation Details}
\label{app:implementation_details}
We build ForceLDM on $\pi_0$~\citep{pi0}, using two camera views and a 10-step history of 6D force/torque measurements as inputs.
Future optical flow is estimated with pretrained RAFT~\citep{Raft} and encoded using SDXL-VAE~\citep{Sdxl}.
We use 16 learnable motion queries and one learnable force query, and align teacher--student representations at layers $\{8,12,16\}$ with a distillation loss weight of $\lambda=0.5$. 
We fully fine-tune a multi-task policy for 30K steps on a single A100-80G GPU with a batch size of 16, using AdamW and a cosine learning-rate schedule that decays from $2.5\times10^{-5}$ to $2.5\times10^{-6}$ after 1,000 warmup steps. At inference, we use 10 flow-matching steps to generate each 32-step action chunk and execute all 32 actions before predicting the next chunk.
Detailed model architecture and training hyperparameters are provided in Appendices~\ref{app:model_architecture} and~\ref{app:hyperparameters}.

\begin{figure}[t]
    \centering
    \includegraphics[width=\linewidth]{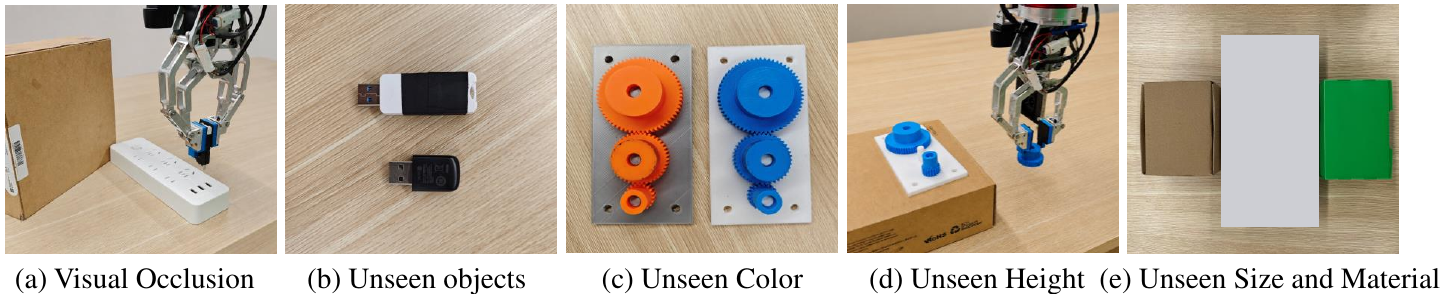}
    \vspace{-5mm}
    \caption{
    Out-of-distribution generalization settings used in our real-world evaluation.
    }
    \label{fig:ood_tasks}
    \vspace{-2mm}
\end{figure}

\begin{table*}[t]
\centering
\caption{Comparison of success rates (\%) on in-distribution (ID) contact-rich manipulation tasks.}
\label{tab:id_results}
\renewcommand{\arraystretch}{1.12}
{\small
\begin{tabular*}{\textwidth}{@{\extracolsep{\fill}}l*{6}{c}@{}}
\toprule
\multicolumn{1}{c}{\textbf{Model}}
& \shortstack{\textbf{USB}\\\textbf{Insertion}}
& \shortstack{\textbf{Gear}\\\textbf{Assembly}}
& \shortstack{\textbf{Box}\\\textbf{Flipping}}
& \shortstack{\textbf{Board}\\\textbf{Wiping}}
& \shortstack{\textbf{VGA}\\\textbf{Plugging}}
& \multicolumn{1}{c@{}}{\textbf{Avg.}} \\
\midrule
\(\boldsymbol{\pi_0}\)~\citep{pi0} w/o force
&  44.0 & 46.7 & 55.0 & 60.0 & 0.0 & 41.1 \\

\addlinespace[1pt]
\(\boldsymbol{\pi_0}\)~\citep{pi0} w/ force
& 64.0 & 83.3 & 70.0 & 70.0 & 5.0 & 58.5 \\

\addlinespace[1pt]
TA-VLA~\citep{TAVLA} 
& 80.0 & \textbf{93.3} & 80.0 & 75.0 & 10.0 & 67.7 \\

\addlinespace[1pt]
ForceVLA~\citep{ForceVLA}
& 52.0 & 66.7 & 65.0 & 65.0 & 5.0 & 50.7 \\

\addlinespace[1pt]
Seer~\citep{Seer} w/ Force

& 76.0 & 70.0 & \textbf{95.0} & 70.0 & 50.0 & 72.2 \\

\addlinespace[1pt]
\textbf{Ours}
& \textbf{92.0} & 83.3 & \textbf{95.0} & \textbf{80.0} & \textbf{85.0} & \textbf{87.1} \\
\bottomrule
\end{tabular*}
}
\end{table*}

\begin{table*}[t]
\centering
\caption{Comparison of success rates (\%) on out-of-distribution (OOD) tasks under visual occlusion and changes in object instance, color, assembly height, material, and size.}
\label{tab:ood_results}
\renewcommand{\arraystretch}{1.12}
{\small
\begin{tabular*}{\textwidth}{@{\extracolsep{\fill}}l*{7}{c}@{}}
\toprule
\multirow[c]{2}{*}{\textbf{Model}}
& \multicolumn{2}{c}{\textbf{USB Insertion}}
& \multicolumn{2}{c}{\textbf{Gear Assembly}}
& \multicolumn{2}{c}{\textbf{Box Flipping}}
& \multirow{2}{*}{\textbf{Avg.}} \\
\cmidrule(lr){2-3}
\cmidrule(lr){4-5}
\cmidrule(lr){6-7}
& Object
& Occlusion
& Height
& Color
& Material
& Size
& \\
\midrule

\addlinespace[1pt]
\(\boldsymbol{\pi_0}\)~\citep{pi0} w/o force
& 20.0 & 26.7
& 0.0 & 40.0
& 26.7 & 0.0
& 18.9 \\

\addlinespace[1pt]
\(\boldsymbol{\pi_0}\)~\citep{pi0} w/ force
& 40.0 & 73.3
& 0.0 & 46.7
& 26.7 & 33.3
& 36.7 \\

\addlinespace[1pt]
TA-VLA~\citep{TAVLA} 
& 20.0 & 66.7
& 13.3 & \textbf{66.7}
& 13.3 & 26.7
& 34.5 \\

\addlinespace[1pt]
ForceVLA~\citep{ForceVLA}
& 26.7 & 20.0
& 0.0 & 40.0
& 0.0 & 13.3
& 16.7 \\

\addlinespace[1pt]
Seer~\citep{Seer} w/ Force
& 46.7 & 20.0
& \textbf{53.3} & 60.0
& 73.3 & 40.0
& 48.9 \\

\addlinespace[1pt]
\textbf{Ours}
& \textbf{86.7} & \textbf{86.7}
& 46.7 & 60.0
& \textbf{80.0} & \textbf{53.3}
& \textbf{68.9} \\
\bottomrule
\end{tabular*}
}
\end{table*}






\begin{table}[t]
\centering
\caption{
Ablation of future modalities used for dynamics distillation.
Visual, Force, and Motion denote future RGB observations, force/torque signals, and optical flow, respectively.
We report success rates (\%) on USB Insertion, Box Flipping, and VGA Plugging.
}
\label{tab:ablation_future_targets}
\renewcommand{\arraystretch}{1.08}
\setlength{\tabcolsep}{4.5pt}
\begin{tabular}{@{}ccc cccc@{}}
\toprule
\multicolumn{3}{c}{\textbf{Future Target}} &
\multicolumn{4}{c}{\textbf{Success Rate (\%)}} \\
\cmidrule(lr){1-3}
\cmidrule(lr){4-7}
Visual & Force & Motion & USB Insertion & Box Flipping & VGA Plugging & Avg. \\
\midrule
--     & --     & \cmark & \textbf{92.0} & 60.0 & 60.0 & 70.7 \\
--     & \cmark & --     & 84.0 & 90.0 & 30.0 & 68.0 \\
\cmark & \cmark & --     & 80.0 & 85.0 & 35.0 & 66.7 \\
--     & \cmark & \cmark & \textbf{92.0} & \textbf{95.0} & \textbf{85.0} & \textbf{90.7} \\
\bottomrule
\end{tabular}
\end{table}

\begin{table}[t]
\centering
\caption{
Ablation of teacher--student distillation components.
$\mathcal{L}_{\mathrm{distill}}$ and $\mathcal{L}_{T}$ denote the representation alignment loss and teacher action loss, respectively. Stop Gradient denotes gradient stopping on teacher targets.
We report success rates (\%) under both in-distribution (ID) and out-of-distribution (OOD) conditions.
}
\label{tab:ablation_architecture}
\renewcommand{\arraystretch}{1.12}
\setlength{\tabcolsep}{5.2pt}
{\small
\begin{tabular*}{0.8\linewidth}{@{\extracolsep{\fill}}lcccccc@{}}
\toprule
\multirow{2}{*}{\textbf{Configuration}}
& \multicolumn{2}{c}{\textbf{USB Insertion}}
& \multicolumn{2}{c}{\textbf{Box Flipping}}
& \multirow{2}{*}{
    \begin{tabular}[c]{@{}c@{}}
    \textbf{VGA}\\
    \textbf{Plugging}
    \end{tabular}
}
& \multirow{2}{*}{\textbf{Avg.}} \\
\cmidrule(lr){2-3}
\cmidrule(lr){4-5}
& \multicolumn{1}{c}{\textbf{ID}}
& \multicolumn{1}{c}{\textbf{OOD}}
& \multicolumn{1}{c}{\textbf{ID}}
& \multicolumn{1}{c}{\textbf{OOD}}
&  &  \\
\midrule
w/o $\mathcal{L}_{\mathrm{distill}}$
& 44.0 & 30.0 & 85.0 & 30.0 & 25.0 & 42.8 \\

w/o $\mathcal{L}_{T}$
& 60.0 & 36.7& 80.0 & 16.7 & 10.0 & 40.7 \\

w/o Stop Gradient
& 68.0 & 50.0 & 60.0 & 16.7 & 20.0 & 42.9 \\

Full Model
& \textbf{92.0} & \textbf{86.7} 
& \textbf{95.0} & \textbf{66.7} 
& \textbf{85.0} 
& \textbf{85.1} \\
\bottomrule
\end{tabular*}
}
\end{table}

\begin{table}[t]
\centering
\caption{
Ablation of the training strategies used in ForceLDM.
Joint Prediction combines action learning with direct future-dynamics prediction, while w/o Progressive Noise removes progressive noise injection into the student's future queries.
}
\label{tab:ablation_training}
\renewcommand{\arraystretch}{1.12}
\setlength{\tabcolsep}{5.2pt}
{\small
\begin{tabular*}{0.8\linewidth}{@{\extracolsep{\fill}}lcccccc@{}}
\toprule
\multirow{2}{*}{\textbf{Training Strategy}}
& \multicolumn{2}{c}{\textbf{USB Insertion}}
& \multicolumn{2}{c}{\textbf{Box Flipping}}
& \multirow{2}{*}{
    \begin{tabular}[c]{@{}c@{}}
    \textbf{VGA}\\
    \textbf{Plugging}
    \end{tabular}
}
& \multirow{2}{*}{\textbf{Avg.}} \\
\cmidrule(lr){2-3}
\cmidrule(lr){4-5}
& \multicolumn{1}{c}{\textbf{ID}}
& \multicolumn{1}{c}{\textbf{OOD}}
& \multicolumn{1}{c}{\textbf{ID}}
& \multicolumn{1}{c}{\textbf{OOD}}
&  &  \\
\midrule
Joint Prediction
& 84.0 & 76.7 & 65.0 & 20.0 & 65.0 & 62.1 \\

w/o Progressive Noise
& 84.0 & 73.3 & 90.0 & 40.0 & \textbf{85.0} & 74.1 \\

Full Model
& \textbf{92.0} & \textbf{86.7}
& \textbf{95.0} & \textbf{66.7}
& \textbf{85.0}
& \textbf{85.1} \\
\bottomrule
\end{tabular*}
}
\end{table}






\subsection{Main Results}
Table~\ref{tab:id_results} summarizes the performance on in-distribution contact-rich manipulation tasks.
ForceLDM achieves the highest average success rate of 87.1\%, outperforming Seer~\citep{Seer}, the strongest baseline, by 14.9 percentage points.
ForceLDM performs particularly well on insertion tasks requiring millimeter-level alignment, achieving 92.0\% on USB Insertion and 85.0\% on VGA Plugging.
We attribute this to learning future contact dynamics, which helps the model understand physical interactions and adjust the insertion pose through contact-based exploration.
In contrast, baselines that model visual dynamics or directly predict future force may overfit to irrelevant details or sensor noise, rather than learning control-relevant information.
Despite these gains, ForceLDM underperforms TA-VLA on Gear Assembly due to limited spatial perception, which sometimes causes grasping failures near the workspace boundary.
Further analysis is provided in Appendix~\ref{app:failure_analysis}.

\subsection{Out-of-Distribution Generalization}
\label{sec:ood_generalization}

\paragraph{OOD Evaluation Settings.}
We evaluate OOD generalization on three representative contact-rich tasks, as shown in Figure~\ref{fig:ood_tasks}.
For USB Insertion, we consider \textbf{Visual Occlusion}, where the external camera view is partially blocked by an object, and \textbf{Unseen Object}, where the USB connector is replaced with a novel instance.
For Gear Assembly, we consider \textbf{Unseen Color}, which changes the object appearance, and \textbf{Unseen Height}, which changes the target assembly height.
For Box Flipping, we consider \textbf{Unseen Material}, where the original wooden box is replaced with a 3D-printed box, leading to different friction properties, and \textbf{Unseen Size}, which evaluates the policy's ability to flip boxes of different sizes.
All OOD settings preserve the original task objectives and success criteria to evaluate robustness to visual and physical changes.
Each OOD setting is evaluated over 15 trials.

\paragraph{Results.}
Table~\ref{tab:ood_results} reports the OOD evaluation results.
ForceLDM achieves the highest average success rate of 68.9\%, outperforming Seer w/ Force, the strongest baseline, by 20.0 percentage points.
Notably, ForceLDM achieves a success rate of 86.7\% on USB Insertion under both unseen objects and visual occlusion, demonstrating strong robustness to changes in object appearance and visual occlusion.
It also achieves the best performance on Box Flipping under material and size changes, showing its ability to handle changes in object properties and contact conditions.
These results suggest that learning future dynamics from optical flow and force signals helps the model focus on physically relevant interactions rather than appearance-specific details, improving generalization across different OOD conditions.

\subsection{Ablation Studies}
\label{sec:ablation}
\paragraph{Ablation on Future Prediction Targets.}
We study which future modalities are most useful for dynamics distillation, as different signals capture different aspects of physical interaction.
We compare four choices: optical flow alone, force alone, force with RGB observations, and force with optical flow.
As shown in Table~\ref{tab:ablation_future_targets}, different modalities provide complementary benefits.
Future force is particularly useful for tasks requiring sustained contact control, achieving 90.0\% success on Box Flipping, but performs poorly on VGA Plugging (30.0\%), where precise motion and alignment are critical.
In contrast, optical flow provides direct motion cues that benefit high-precision manipulation tasks, achieving 92.0\% on USB Insertion, but performs less effectively on tasks requiring sustained force control, such as Box Flipping.
Adding future RGB observations to force does not improve performance, possibly because the model focuses on irrelevant visual details rather than control-relevant interaction dynamics.
Combining force and optical flow achieves the highest average success rate of 90.7\%, substantially outperforming force alone at 68.0\% and optical flow alone at 70.7\%.
These results support our choice of future force and optical flow as complementary supervision signals for dynamics distillation.

\textbf{Ablation on Model Architecture.}
We study the importance of key components in our teacher--student architecture for learning action-relevant future dynamics.
We consider three variants by removing the distillation loss $\mathcal{L}_{\mathrm{distill}}$, teacher action loss $\mathcal{L}_{T}$, and stop-gradient operation, respectively.
As shown in Table~\ref{tab:ablation_architecture}, removing $\mathcal{L}_{\mathrm{distill}}$ reduces the average success rate from 85.1\% to 42.8\%, indicating that action supervision alone is insufficient to transfer future dynamics from the teacher to the student.
Removing $\mathcal{L}_{T}$ also leads to a substantial performance drop, suggesting that the teacher must be grounded in action prediction to learn useful representations for distillation.
Similarly, removing stop-gradient leads to representation collapse, resulting in a substantial performance drop.
These results validate the effectiveness of our teacher--student architecture and its key design choices.

\textbf{Ablation on Training Strategies.}
Table~\ref{tab:ablation_training} studies the importance of latent dynamics distillation and progressive noise injection in ForceLDM.
First, we compare latent dynamics distillation with an alternative strategy that directly predicts future signals through joint prediction.
This variant achieves an average success rate of only 62.1\%, substantially below the full model.
This suggests that direct prediction may encourage the model to focus on irrelevant details and sensor noise rather than learning control-relevant dynamics.
Second, removing progressive noise reduces the average success rate from 85.1\% to 74.1\%.
The performance drop is larger in OOD settings than in ID settings, suggesting that progressive noise injection helps improve the model's generalization ability.

\vspace{-2mm}
\subsection{Visualization}
\label{sec:subsection}
\begin{figure}[t]
    \centering
    \setlength{\abovecaptionskip}{3pt}
    \includegraphics[width=\linewidth]{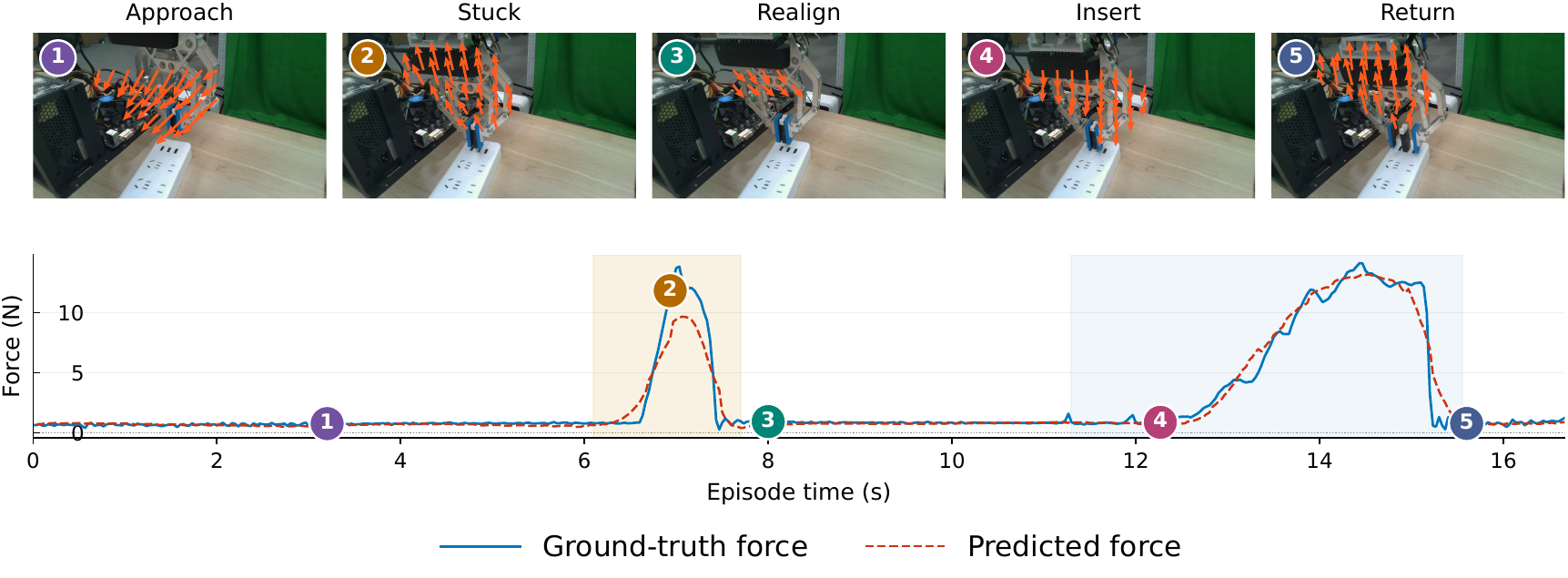}
    \caption{\textbf{Probing Latent Dynamics.}
    We freeze the trained model and fit lightweight probes to predict optical flow and contact forces from its intermediate hidden states.
    \textbf{Top:} Predicted future 2D motion vectors derived from optical flow (bright orange-red arrows) at five stages of the USB insertion task.
    \textbf{Bottom:} Measured contact force (blue solid) and predicted force (red dashed).
    }
    \label{fig:usb_retry_probe}
    \vspace{-5mm}
\end{figure}

\textbf{Visualization of Latent Dynamics.}
To examine whether the learnable future queries capture future motion and contact dynamics, we train lightweight probes to predict future optical flow and contact forces from the future query representations at layer 12 of the frozen policy.
Figure~\ref{fig:usb_retry_probe} shows five stages of a USB insertion task: approach, stuck, realignment, insertion, and return.
When the gripper becomes stuck, the predicted optical flow indicates an upward motion for recovery.
During realignment, the predicted flow shows the direction of the next adjustment, followed by downward motion during insertion.
The predicted force follows the overall trend of the measured force, capturing the main changes during two contact events. Although the predicted values differ from the measurements, the predicted force is smoother and does not reproduce high-frequency noise in the measured signal.
These results suggest that the learned representations capture future motion and contact dynamics while avoiding overfitting to irrelevant details.

\textbf{Additional Visualizations and Failure Case Analysis.}
Additional visualizations of successful task executions under ID and OOD conditions are provided in Appendix~\ref{app:visualization}.
Representative failure cases and an analysis of common failure modes are presented in Appendix~\ref{app:failure_analysis}.

\section{Conclusion}
We propose \textbf{ForceLDM}, a force-aware latent dynamics model for contact-rich manipulation.
At its core, ForceLDM distills future force/torque and optical-flow cues from a privileged teacher into a student policy, enabling it to capture action-relevant interaction dynamics using only current observations at inference.
Experiments on five real-world contact-rich manipulation tasks demonstrate that ForceLDM outperforms force-aware VLA baselines in both task performance and generalization to unseen visual and physical variations.
Ablation studies validate the complementary roles of future force and optical flow, together with the benefits of progressive noise injection.
These results highlight the value of learning future physical dynamics for force-aware robotic manipulation.

\textbf{Limitations.}
\label{sec:limitations}
Despite the promising results of ForceLDM, several limitations remain.
First, our experiments use small-scale datasets, and whether ForceLDM maintains its advantage as training data scales up remains unclear.
Second, optical-flow estimation errors caused by occlusion or motion blur may affect the learned dynamics representations.
Third, 2D optical flow lacks explicit depth information, which may limit the model's ability to capture spatial relationships during manipulation.
Future work will explore scaling up training data to improve performance and combining 3D scene flow with force feedback to better model future interaction dynamics.

\vspace{-2mm}
\section*{Acknowledgments}
\vspace{-2mm}
This work was supported by the National Natural Science Foundation of China (NSFC) under Grant Nos. 62402367, 62232008, and 2024CSJGG01500, the Fundamental Research Funds for the Central Universities, and the Jiangsu Funding Program for Excellent Postdoctoral Talent under Grant No.2025ZB315. The authors declare no competing interests.


\bibliographystyle{plainnat}
\bibliography{references}

\appendix


\newpage


\section{Detailed Experimental Setup}
\label{sec:data_collection}

\begin{figure}[!h]
    \centering
    \begin{minipage}[t]{0.48\linewidth}
        \centering
        \includegraphics[width=\linewidth,height=4.5cm,keepaspectratio]{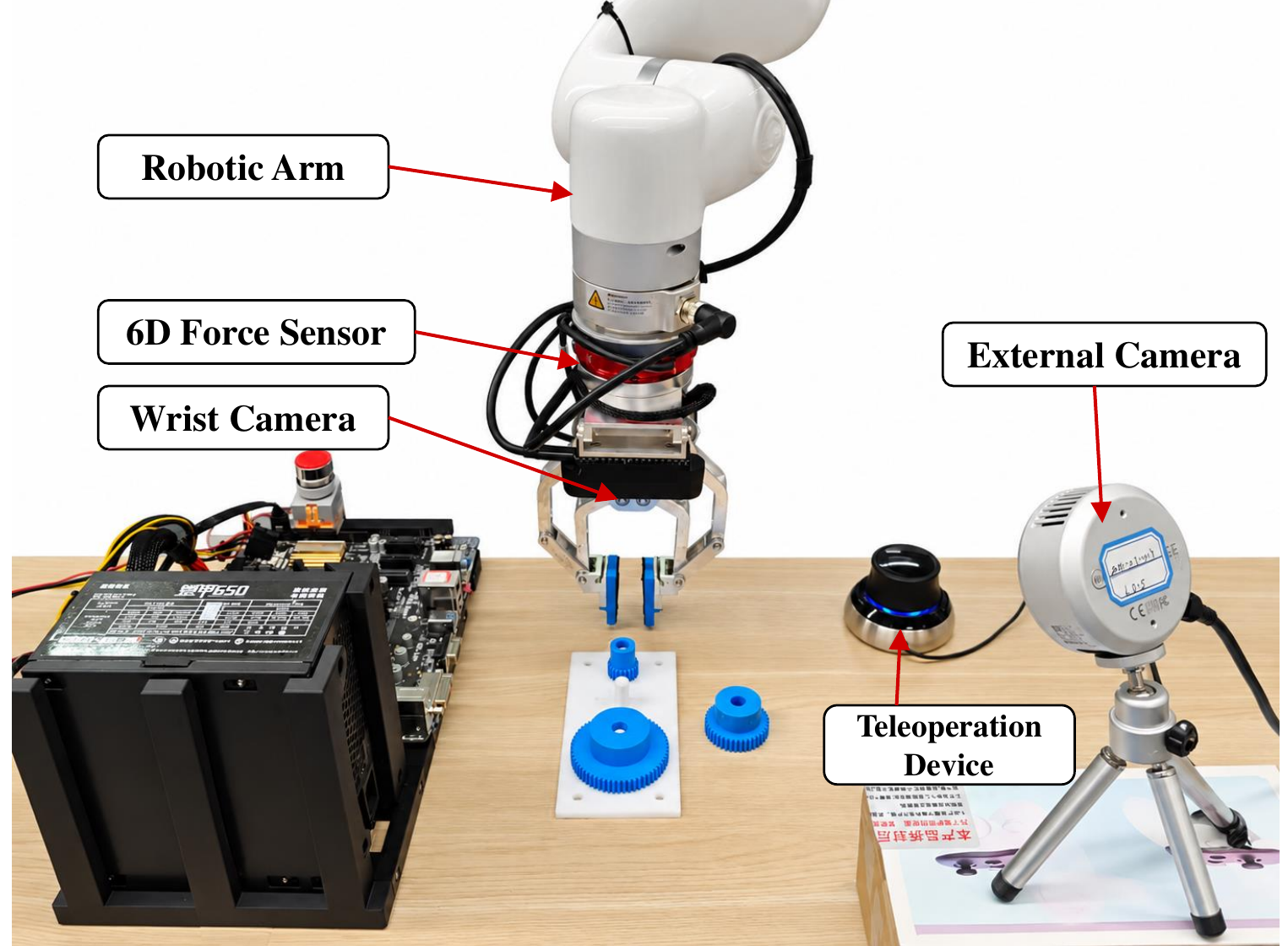}
        \par\smallskip
        (a) xArm 7
    \end{minipage}\hfill
    \begin{minipage}[t]{0.48\linewidth}
        \centering
        \includegraphics[width=\linewidth,height=4.5cm,keepaspectratio]{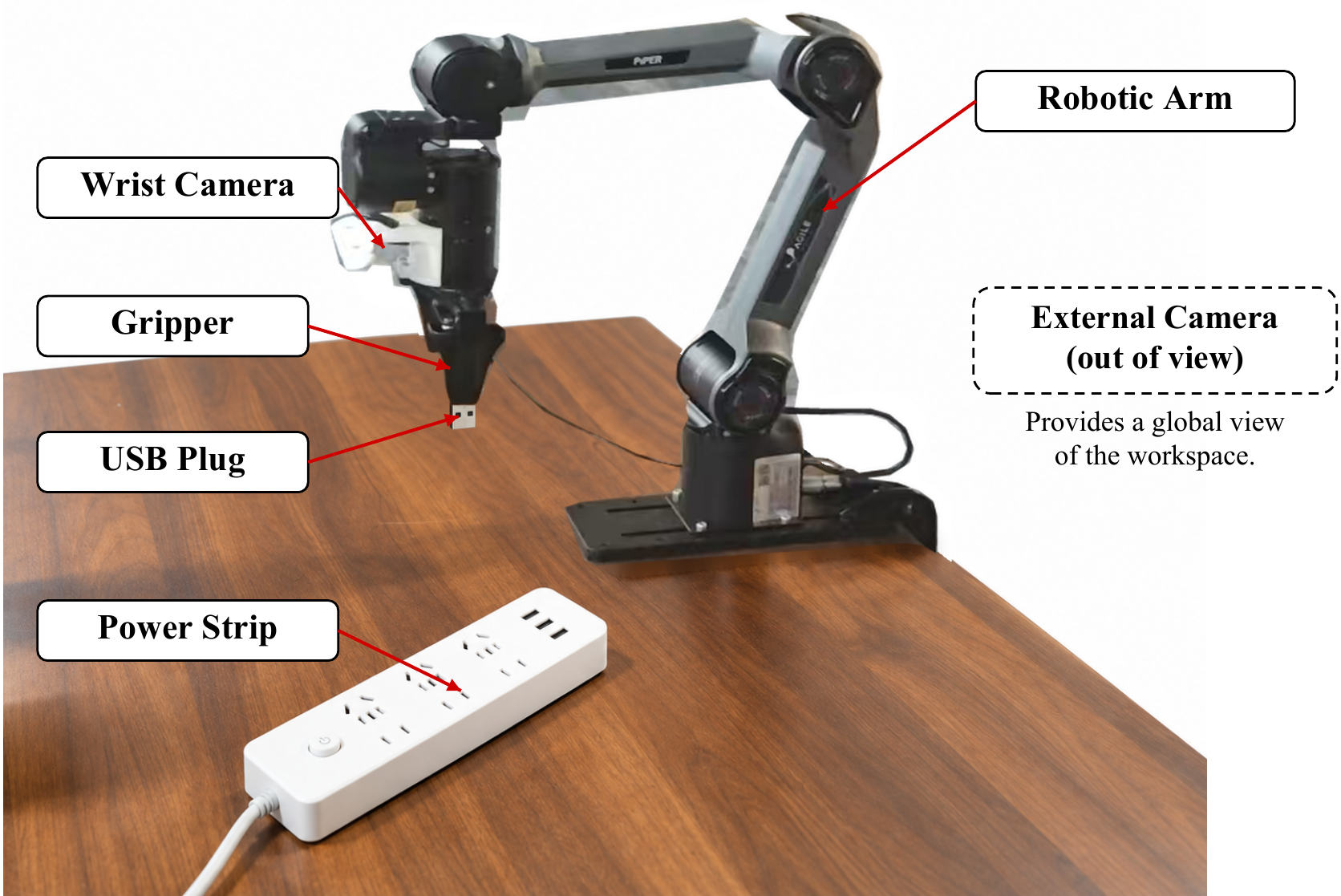}
        \par\smallskip
        (b) Piper-X
    \end{minipage}
    \caption{Real-world experimental setups for (a) the main xArm 7 experiments and (b) the supplementary Piper-X evaluation.}
    \label{fig:appendix_setup}

\end{figure}

\subsection{Robot Platform} 
We conduct most experiments on xArm 7 and further evaluate our method on Piper-X with a different system configuration, as shown in Figure~\ref{fig:appendix_setup}.
The main setup uses a 7-DoF xArm 7 equipped with a wrist-mounted 6D force/torque sensor, a third-person RealSense L515 camera, and an Orbbec Gemini 336L wrist camera, using $\pi_0$ as the policy backbone.
The supplementary setup uses a 6-DoF AgileX Piper-X with joint-torque estimates derived from motor currents, a RealSense D435 camera, and an Orbbec Gemini 335 camera, with GR00T N1.5 as the policy backbone.
Table~\ref{tab:piper_platform} compares their configurations.

\begin{table}[!h]
    \centering
    \caption{Comparison of experimental setups across robot platforms.}
    \label{tab:piper_platform}
    {\small
    \renewcommand{\arraystretch}{1.12}
    \setlength{\tabcolsep}{6pt}
    \begin{tabular*}{\linewidth}{@{\extracolsep{\fill}}ccc@{}}
    \toprule
    \textbf{Component} & \textbf{xArm 7} & \textbf{Piper-X} \\
    \midrule
    DoF & 7 & 6 \\
    Force input & 6D F/T sensor & \shortstack{Joint torque} \\
    Cameras & RealSense L515 + Gemini 336L & RealSense D435 + Gemini 335 \\
    Policy backbone & \(\pi_0\) & GR00T N1.5 \\
    \bottomrule
    \end{tabular*}
    }
\end{table}

\subsection{Data Collection}
\label{app:piper_data_collection}
We collect demonstrations through 3D SpaceMouse teleoperation, recording synchronized camera observations, robot states, force/torque measurements, and actions. To improve data diversity, we vary the initial robot pose and include demonstrations that recover from failed attempts. For example, in USB insertion, the operator deliberately misaligns the plug with the port and then demonstrates corrective motions to complete the insertion, instead of always starting from an aligned pose. Approximately 20--30\% of the demonstrations include such recovery behaviors. For each robot platform, we combine demonstrations from all tasks into a single multi-task dataset for policy training. Dataset statistics for both robot platforms are summarized in Table~\ref{tab:dataset_summary}.

\begin{table}[h]
\centering
\caption{Dataset statistics for the xArm 7 and Piper-X robot platforms.}
\label{tab:dataset_summary}
\small
\renewcommand{\arraystretch}{1.15}
\setlength{\tabcolsep}{8pt}
\begin{tabular*}{\linewidth}{@{\extracolsep{\fill}}clccc@{}}
\toprule
\textbf{Robot Platform} & \textbf{Task} & \textbf{Demos} & \textbf{Frames} & \textbf{Duration (h)} \\
\midrule
\multirow{6}{*}{xArm 7} & USB Insertion & 100 & 55,667 & 0.52 \\
& Gear Assembly & 80 & 64,091 & 0.59 \\
& Box Flipping & 80 & 71,795 & 0.66 \\
& Board Wiping & 80 & 70,810 & 0.66 \\
& VGA Plugging & 80 & 69,874 & 0.65 \\
\cmidrule{2-5}
& \textbf{Total} & \textbf{420} & \textbf{332,237} & \textbf{3.08} \\
\midrule
\multirow{3}{*}{Piper-X} & USB Insertion & 100 & 48,438 & 0.45 \\
& Charger Insertion & 100 & 41,651 & 0.39 \\
\cmidrule{2-5}
& \textbf{Total} & \textbf{200} & \textbf{90,089} & \textbf{0.83} \\
\bottomrule
\end{tabular*}
\end{table}

\subsection{Evaluation Protocol}
\label{app:evaluation_details}
For each task, the robot starts from a fixed initial pose, while the object is placed at one of a predefined set of evaluation locations. Each method is evaluated on the same set of object locations with the same number of trials. For the main in-distribution evaluation on xArm 7, each method is evaluated over 25 trials on USB Insertion, 30 trials on Gear Assembly, and 20 trials each on Box Flipping, Board Wiping, and VGA Plugging. A trial is considered successful only if the task is completed within 2 minutes and the contact force remains below the safety threshold of 25\,N. Otherwise, the trial is counted as a failure. The success rate is calculated as the number of successful trials divided by the total number of evaluation trials.

\subsection{Task Descriptions} 
We describe each task and its key challenges below, with representative successful rollouts provided in Appendix~\ref{app:visualization}.

\begin{itemize}
    \item \textbf{USB Insertion.}
    The robot must align a USB plug with the target port on a power strip and insert it while adjusting its position and orientation. This task requires precise alignment due to the tight clearance between the plug and the port. The task is successful when the plug is fully inserted into the port.

    \item \textbf{Gear Assembly.}
    The robot must pick up a gear, align its center hole with the target shaft, and move it downward to complete the assembly. Since the hole is occluded during alignment, the robot must rely on force feedback to adjust its position. The task is successful when the gear is fully mounted on the shaft.

    \item \textbf{Box Flipping.}
    The robot must flip a box from a flat position to an upright position using a fixed support. During flipping, the robot must continuously adjust its motion based on force feedback to prevent the box from slipping or excessive force from being applied. The task is successful when the box stands upright on its narrow face.

    \item \textbf{Board Wiping.}
    The robot must pick up an eraser and use it to wipe the marked region of a board. During wiping, the robot must maintain sufficient contact force to effectively remove the markings. The task is successful when all markings in the target region are removed.

    \item \textbf{VGA Plugging.}
    The robot must pick up a VGA connector, align it with the target socket, and insert it horizontally. The irregular shape of the connector requires precise alignment and force-guided position adjustments during insertion. The task is successful when the connector is fully inserted into the socket.
\end{itemize}

\section{Model Architecture Details}
\label{app:model_architecture}

\textbf{Backbone.}
We use the pretrained PaliGemma backbone from $\pi_0$~\citep{pi0}, comprising a SigLIP image encoder and a Gemma-2B language model with approximately 2.92B parameters in total.
It takes two RGB images from the external and wrist cameras, each resized with padding to $224\times224$, together with the task instruction.
The image and instruction tokens form a shared visual--language prefix for both action experts.

\textbf{Teacher and Student Experts.}
Both experts use the same approximately 300M-parameter architecture with 18 Transformer layers and hidden width 1024.
Their Transformer weights are initialized from the same pretrained $\pi_0$ action expert and then optimized separately within joint training.
Both receive the shared prefix, robot state, and force history, and predict 32-step action chunks through flow matching.
The teacher additionally receives encoded future force and motion tokens, while the student uses learnable future queries.

\textbf{Force Encoding.}
We flatten ten historical 6D force/torque measurements and use a one-layer MLP to produce one  force token for each expert.
A separate one-layer MLP encodes the teacher's 32-step future force sequence into one token of the same width.

\textbf{Future Motion Encoding.}
RAFT~\citep{Raft} estimates optical flow between the current and future images for both views.
We convert each flow field to RGB by mapping direction to hue and using normalized, clipped magnitude to control color strength against a white background.
A frozen SDXL-VAE~\citep{Sdxl} encodes these motion images.
After $2\times2$ latent patch merging, a two-layer MLP and layer normalization project the spatial features to width 1024.
A Q-Former-style compressor with 16 learnable queries aggregates the features from both views through cross-attention, producing 16 motion tokens.

\textbf{Student Future Queries.}
The student uses 17 learnable queries of width 1024: one for future force and 16 for future motion, matching the teacher's privileged tokens.
Each expert concatenates its force-history token, state token, future-related tokens, and noisy action tokens.
Future queries attend to causal observations but not to noisy action tokens.

\textbf{Representation Alignment.}
We align future-related hidden states at layers $\{8,12,16\}$ using separate two-layer force and motion projectors with Swish activations.
The projected student states are matched to stop-gradient teacher targets using cosine distance.

\section{Training Hyperparameters}
\label{app:hyperparameters}
ForceLDM contains approximately 3.56B trainable parameters, including a 2.92B vision--language backbone and two 311M action experts. We jointly fine-tune the backbone, both action experts, and the additional modules, while keeping the RAFT and SDXL-VAE preprocessing models frozen. The training hyperparameters are summarized in Table~\ref{tab:training_hyperparameters}.

\begin{table}[htbp]
\centering
\caption{Training hyperparameters for ForceLDM.}
\label{tab:training_hyperparameters}
\small
\renewcommand{\arraystretch}{1.06}
\setlength{\tabcolsep}{6pt}
\begin{tabular}{p{0.53\linewidth}p{0.39\linewidth}}
\toprule
\textbf{Parameter} & \textbf{Value} \\
\midrule
\multicolumn{2}{l}{\textbf{Optimization}} \\
Training hardware & 1 NVIDIA A100, 80 GB \\
Training steps & 30,000 \\
Batch size & 16 \\
Optimizer & AdamW \\
Peak learning rate & $2.5\times10^{-5}$ \\
Final learning rate & $2.5\times10^{-6}$ \\
Learning-rate schedule & Cosine decay \\
Learning-rate warmup & 1,000 steps \\
Adam $(\beta_1,\beta_2)$ & $(0.9,0.95)$ \\
Gradient clipping (global norm) & 1.0 \\
\midrule
\multicolumn{2}{l}{\textbf{Distillation}} \\
Alignment layers & $\{8,12,16\}$ \\
Distillation weight $\lambda$ & 0.5 \\
Motion / force queries & 16 / 1 \\
\midrule
\multicolumn{2}{l}{\textbf{Query Perturbation}} \\
Maximum query noise scale $\sigma_{\max}$ & $0.3$ \\
Noise ramp start $p_{\mathrm{start}}$ & $0.3$ \\
Noise ramp end $p_{\mathrm{end}}$ & $0.7$ \\
\bottomrule
\end{tabular}
\end{table}

\section{Baseline Implementation Details}
\label{app:baseline_details}
We implement all baselines except ForceVLA within our $\pi_0$-based framework, following their official open-source implementations. All methods are trained on the same task datasets for 30,000 optimization steps with a batch size of 16, using the optimizer settings in Table~\ref{tab:training_hyperparameters}. For ForceVLA, we use its official codebase to preserve the original fusion architecture.

\paragraph{$\boldsymbol{\pi_0}$ w/o Force.}
The base $\pi_0$~\citep{pi0} combines a PaliGemma vision--language backbone with a Transformer action expert. The backbone encodes the two camera views and language instruction, while the action expert processes robot-state and noisy-action tokens conditioned on the vision--language context. It predicts 32-step action chunks through flow matching without force input.

\paragraph{$\boldsymbol{\pi_0}$ w/ Force.}
We retain the original backbone, action expert, and flow-matching objective, while introducing an MLP to encode the force history into a single token. This force token is concatenated with the state and noisy-action tokens as input to the action expert. The model thus incorporates force feedback without explicitly modeling future interaction dynamics.

\paragraph{TA-VLA.}
Following TA-VLA~\citep{TAVLA}, we encode the historical force sequence using a two-layer MLP and feed the resulting force token into the action expert alongside the state and noisy-action tokens. Unlike $\pi_0$ w/ Force, this baseline extends the flow-matching objective to jointly predict future action chunks and force sequences. The additional force-prediction loss encourages the model to learn force-related information during action generation. We set the auxiliary force-prediction loss weight to 0.1.

\paragraph{ForceVLA.}
ForceVLA~\citep{ForceVLA} extends $\pi_0$ with FVLMoE, a force-aware Mixture-of-Experts fusion module that dynamically integrates pretrained vision--language features with real-time 6-axis force feedback during action decoding. The module uses modality-specific experts to adaptively fuse multimodal information, enabling force-aware action prediction. We train ForceVLA on our demonstrations using its official implementation.

\paragraph{Seer w/ Force.}
Seer~\citep{Seer} introduces visual foresight to guide action prediction through future visual representations. We adapt this design to $\pi_0$ by introducing ten learnable foresight tokens per camera view to extract future visual features from the vision--language backbone. These features are directly provided to the action expert for flow-matching-based action prediction, while a separate masked image decoder reconstructs future RGB images under an MSE loss. We additionally incorporate historical force observations and jointly optimize action prediction and future RGB reconstruction. The force history is encoded by a two-layer MLP into a single token, which is concatenated with the state and noisy-action tokens as input to the action expert. We set the RGB reconstruction loss weight to 0.1.

\section{Additional Experiments}
\label{app:additional_experiments}

\subsection{Evaluation on Another Robot Platform}
\label{app:piper_gr00t}
To examine whether our method can be applied to different robot platforms and VLA architectures, we conduct additional real-world experiments on an alternative robotic platform. Compared with our main experiments using xArm 7 and $\pi_0$, these experiments employ a different robot, force-sensing method, camera configuration, and VLA backbone. 

\paragraph{Experimental Setup.}
We use an AgileX Piper-X manipulator and adopt GR00T N1.5~\citep{GR00TN15} as the VLA policy backbone.
The force-sensing and camera configurations are described in Appendix~\ref{sec:data_collection} and compared with the main xArm 7 setup in Table~\ref{tab:piper_platform}.
To implement ForceLDM on this backbone, we duplicate the policy's action-decoding head to construct the teacher and student experts, which are trained using the same future-dynamics distillation framework as in the main experiments.
We consider two tasks: inserting a USB plug and inserting a charger plug into their corresponding sockets on a power strip.
We use the separate Piper-X dataset described in Appendix~\ref{app:piper_data_collection}, with 100 demonstrations per task (200 in total), for platform-specific policy adaptation. We evaluate each method over 20 trials per task.
All compared methods use the same task initialization and evaluation protocol.

\paragraph{Results.}
Table~\ref{tab:piper_results} presents the success rates on two real-world manipulation tasks using Piper-X. ForceLDM achieves 80\% on USB insertion and 60\% on charger insertion, averaging 70\% and outperforming GR00T N1.5 and TA-VLA by 50 and 7.5 percentage points, respectively. In contrast, directly adding torque observations reduces GR00T N1.5's average success rate to 17.5\%, potentially due to noisy torque signals and the dominance of visual features. By modeling action-relevant future interaction dynamics, ForceLDM better leverages force feedback for precise manipulation. These results demonstrate the effectiveness of our method on a different robot platform and VLA backbone.

\begin{table}[htbp]
\centering
\caption{Real-world manipulation success rates (\%) on the Piper-X platform. Each method is evaluated over 20 trials per task, with the average success rate reported across both tasks.}
\label{tab:piper_results}
\small
\renewcommand{\arraystretch}{1.15}
\setlength{\tabcolsep}{10pt}
\begin{tabular}{lccc}
\toprule
\textbf{Method} & \textbf{USB Insertion} & \textbf{Charger Insertion} & \textbf{Avg.} \\
\midrule
GR00T N1.5 & 15.0 & 25.0 & 20.0 \\
GR00T N1.5 + Torque Input & 25.0 & 10.0 & 17.5 \\
TA-VLA & 70.0 & 55.0 & 62.5 \\
\textbf{ForceLDM (Ours)} & \textbf{80.0} & \textbf{60.0} & \textbf{70.0} \\
\bottomrule
\end{tabular}
\end{table}

\subsection{Sensitivity to the Prediction Horizon}
\label{app:horizon_ablation}
The prediction horizon determines how far into the future ForceLDM learns to model interaction dynamics.
To investigate its effect, we evaluate three horizons \(H\in\{16,32,50\}\) on Box Flipping and Gear Assembly under identical settings.
Specifically, \(H\) controls both the action-chunk length and the future-prediction horizon, where the model predicts \(H\) future force steps and optical flow between \(I_t\) and \(I_{t+H}\).
As shown in Table~\ref{tab:horizon_ablation}, \(H=16\) and \(H=32\) achieve comparable performance, while increasing the horizon to \(H=50\) substantially reduces success rates.
This suggests that excessively long horizons may introduce greater uncertainty in future interaction dynamics and make precise contact control more challenging.
We therefore adopt \(H=32\), which achieves the highest average success rate.

\begin{table}[htbp]
\centering
\caption{Effect of the prediction horizon on task success rates (\%).}
\small
\label{tab:horizon_ablation}
\renewcommand{\arraystretch}{1.15}
\setlength{\tabcolsep}{12pt}
\begin{tabular}{cccc}
\toprule
\textbf{Horizon \(H\)} & \textbf{Box Flipping} & \textbf{Gear Assembly} & \textbf{Average} \\
\midrule
16 & 100.0 & 86.7 & 93.4 \\
\textbf{32} & \textbf{100.0} & \textbf{90.0} & \textbf{95.0} \\
50 & 90.0 & 70.0 & 80.0 \\
\bottomrule
\end{tabular}
\end{table}

\subsection{Sensitivity to Optical-Flow Quality}
\label{app:flow_estimator}
ForceLDM learns future interaction dynamics from estimated optical flow, whose quality may vary across different estimators.
To examine whether our method relies on a specific flow estimator, we compare RAFT (default) and DPFlow~\citep{DPFlow} on USB and charger insertion using the Piper-X and GR00T N1.5 setup described in Appendix~\ref{app:piper_gr00t}, while keeping all other settings fixed.
We include GR00T N1.5 and TA-VLA as baselines, using the results reported in Table~\ref{tab:piper_results}.
As shown in Table~\ref{tab:flow_estimator}, replacing RAFT with DPFlow changes the success rate from 80\% to 75\% on USB insertion and from 60\% to 70\% on charger insertion.
The comparable performance suggests that our method is not strongly dependent on the choice of flow estimator, as both provide useful motion cues for learning future interaction dynamics.

\begin{table}[htbp]
\centering
\caption{Comparison of optical-flow estimators on real-world manipulation success rates (\%) using Piper-X.}
\label{tab:flow_estimator}
\small
\renewcommand{\arraystretch}{1.15}
\setlength{\tabcolsep}{10pt}
\begin{tabular}{lccc}
\toprule
\textbf{Configuration} & \textbf{USB Insertion} & \textbf{Charger Insertion} & \textbf{Avg.} \\
\midrule
GR00T N1.5 & 15.0 & 25.0 & 20.0 \\
TA-VLA & 70.0 & 55.0 & 62.5 \\
\midrule
ForceLDM (RAFT) & \textbf{80.0} & 60.0 & 70.0 \\
ForceLDM (DPFlow) & 75.0 & \textbf{70.0} & \textbf{72.5} \\
\bottomrule
\end{tabular}
\end{table}




\section{Visualization}
\label{app:visualization}
We provide representative successful executions of ForceLDM under both in-distribution (ID) and out-of-distribution (OOD) conditions.

\begin{figure}[p]
    \centering
    \includegraphics[width=\linewidth,height=0.88\textheight,keepaspectratio]{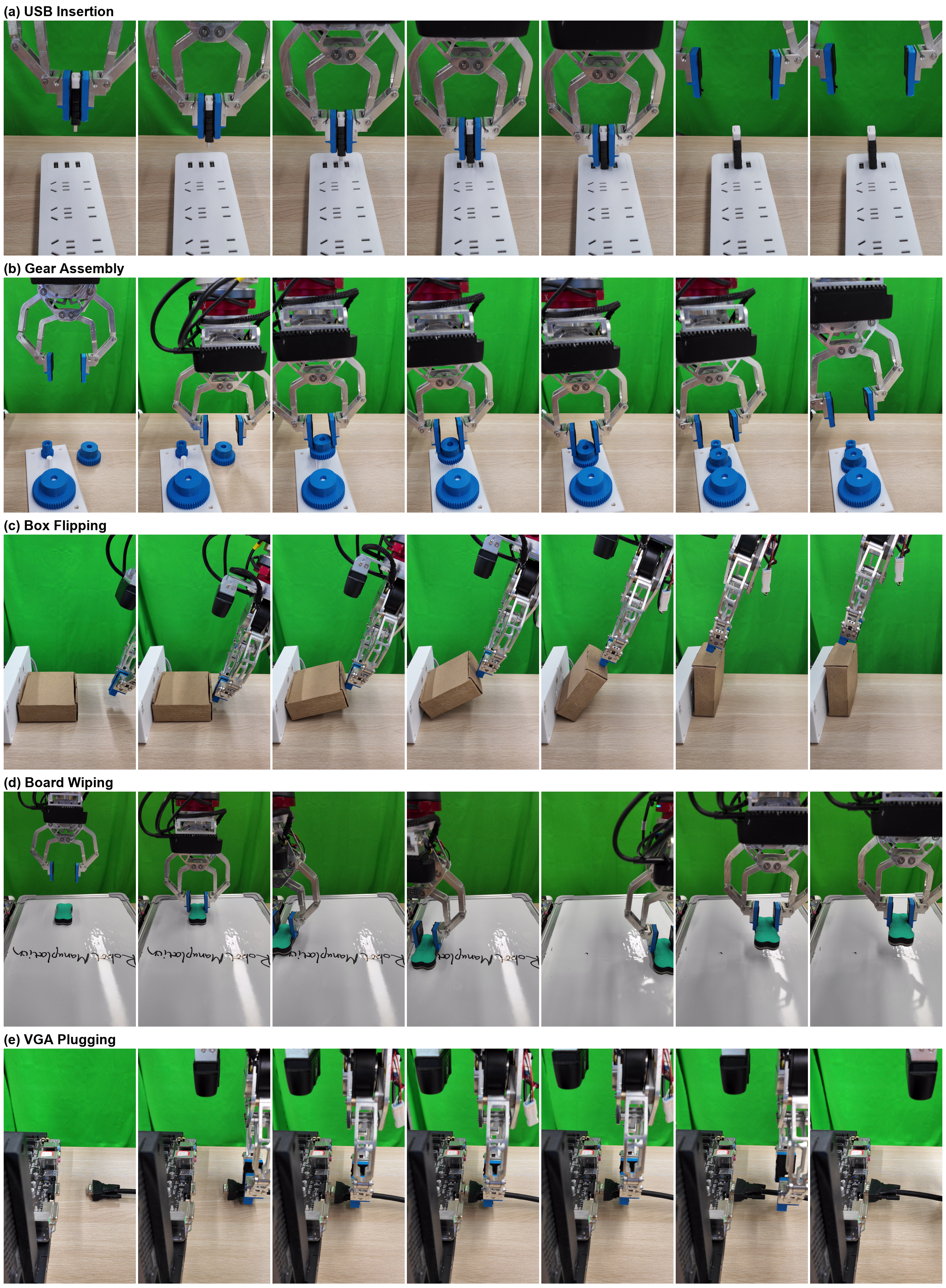}
    \caption{Successful executions of five real-world manipulation tasks: USB Insertion, Gear Assembly, Box Flipping, Board Wiping, and VGA Plugging. Each row shows seven frames from the initial state to task completion.}
    \label{fig:task_execution}
\end{figure}

Figure~\ref{fig:task_execution} illustrates successful executions of five real-world manipulation tasks. These tasks involve different forms of contact-rich interactions, including precise insertion, gear assembly, object flipping, and surface wiping. ForceLDM successfully completes these tasks by handling contact interactions that require precise alignment and force control. In particular, the insertion and assembly tasks require accurate positioning under tight clearances, while flipping and wiping require appropriate contact forces throughout the manipulation process. These examples highlight the effectiveness of our method in contact-rich manipulation.

\begin{figure}[p]
    \centering
    \includegraphics[width=\linewidth,height=0.88\textheight,keepaspectratio]{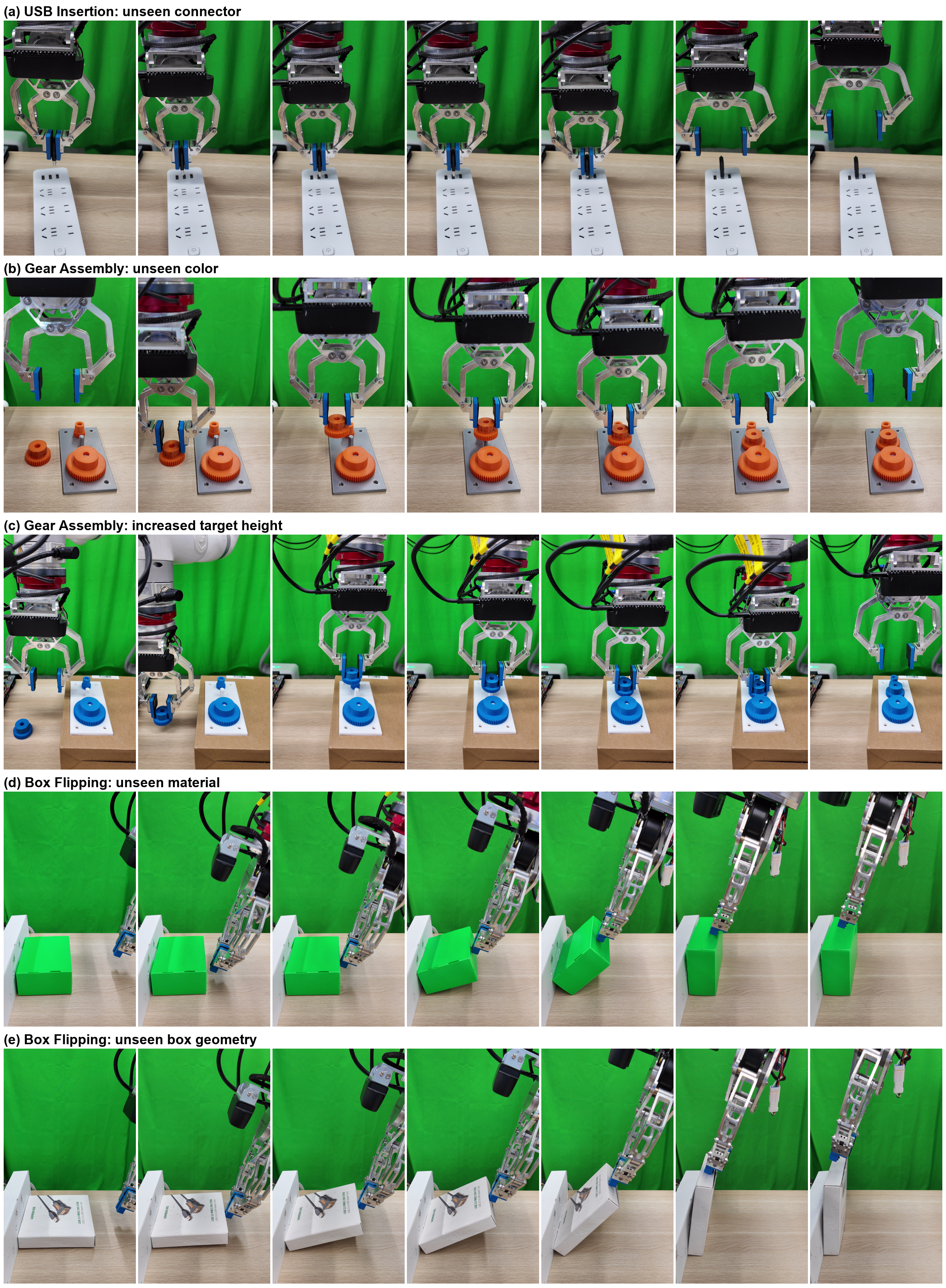}
    \caption{Successful executions under OOD conditions: (a) an unseen USB connector, (b) an unseen gear color, (c) an increased assembly height, (d) an unseen box material, and (e) an unseen box size. Each row shows seven frames from the initial state to task completion.}
    \label{fig:ood_rollout_sequences}
\end{figure}

Figure~\ref{fig:ood_rollout_sequences} further presents successful executions under unseen conditions, including changes in object appearance, material, size, and assembly height. Despite these variations, ForceLDM successfully completes the tasks, demonstrating its ability to handle different object properties and spatial configurations. These results further suggest that learning action-relevant future interaction dynamics helps the policy maintain effective manipulation performance beyond the training conditions.

\section{Failure Case Analysis}
\label{app:failure_analysis}
\begin{figure}[p]
    \centering
    \includegraphics[width=\linewidth,height=0.88\textheight,keepaspectratio]{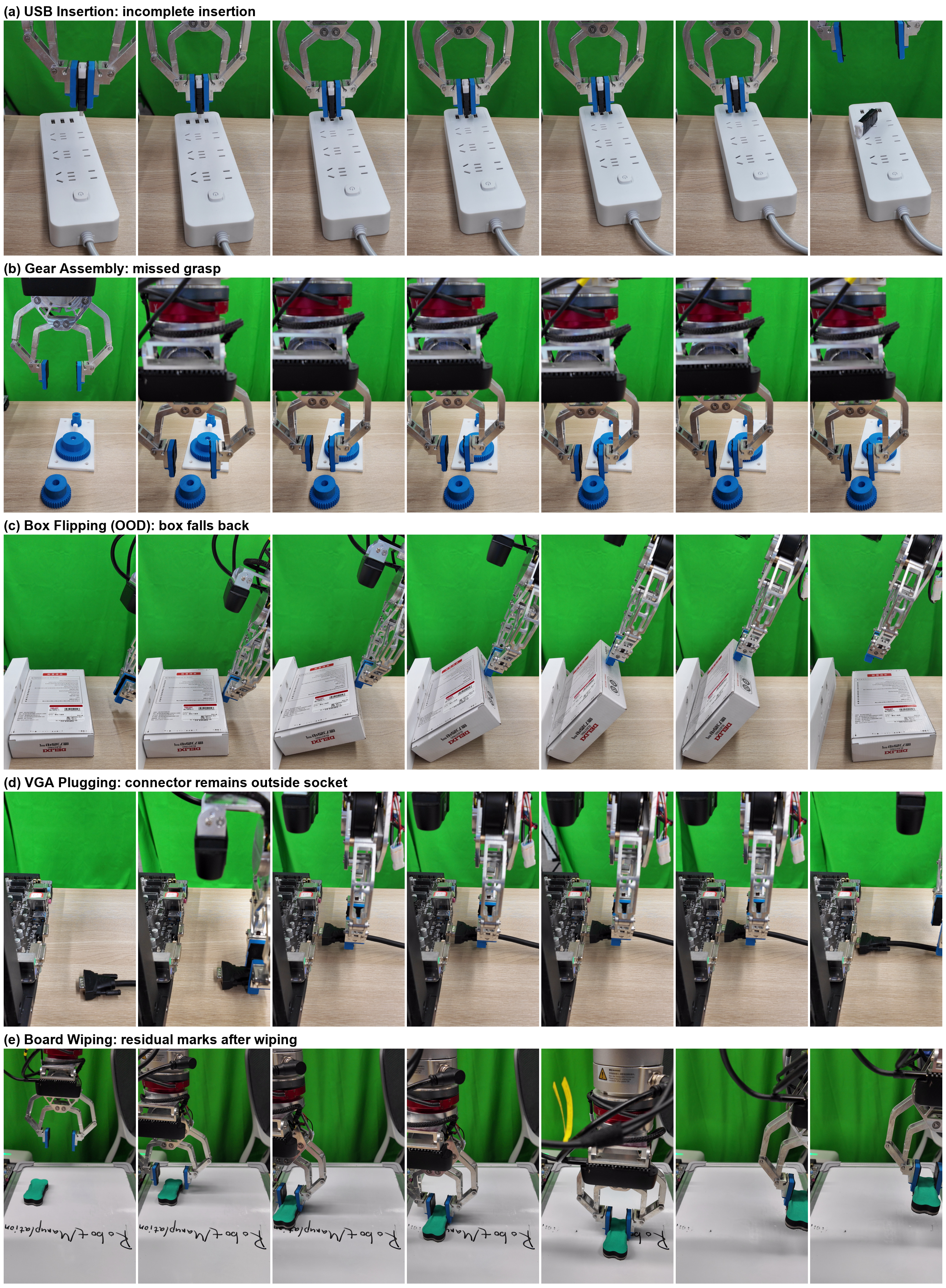}
    \caption{Representative failure cases: (a) the USB plug failing to fully insert into the target port, (b) a missed gear grasp, (c) a box falling back during flipping under OOD conditions, (d) the VGA connector released before being fully inserted, and (e) small residual marks after wiping. Each row shows seven frames from the initial state to the terminal failure state. }
    \label{fig:failure_case_sequences}
\end{figure}

Figure~\ref{fig:failure_case_sequences} presents representative failure cases from our real-world experiments. We identify three common failure modes.

\begin{itemize}
    \item \textbf{Repeated Attempts.}
    In USB Insertion and VGA Plugging (Figure~\ref{fig:failure_case_sequences}(a,d)), the robot may repeatedly attempt insertion without successfully completing the task. In some cases, the robot releases the connector before it is fully inserted, resulting in task failure.

    \item \textbf{Unstable or Insufficient Contact.}
    In Box Flipping (Figure~\ref{fig:failure_case_sequences}(c)), insufficient contact force may cause the box to slip during flipping. Similarly, in Board Wiping (Figure~\ref{fig:failure_case_sequences}(e)), the robot may fail to maintain sufficient force to remove all markings from the board.

    \item \textbf{Grasping Failures.}
    In Gear Assembly (Figure~\ref{fig:failure_case_sequences}(b)), the robot occasionally closes its gripper before reaching the gear, resulting in a failed grasp. This may arise from inaccurate estimation of the distance between the gripper and the gear due to limited depth information.
\end{itemize}

Similar failure modes are also observed in the baseline methods, indicating that precise contact control and reliable grasping remain challenging in real-world manipulation.

\clearpage

\end{document}